\documentclass{article}
\usepackage {url}
\usepackage[utf8]{inputenc}
\usepackage{algorithm}
\usepackage{algorithmic}
\usepackage{amsmath}
\usepackage{multirow}
\usepackage{graphicx}
\usepackage{epstopdf}
\usepackage{float}
\usepackage{rotating}
\usepackage{graphicx}
\usepackage{hyperref}
\usepackage{caption}
\usepackage{subcaption}
\let\captionbox\cpbox
\usepackage{geometry}
\usepackage{changepage}
\usepackage[accepted]{icml2013}
\let\cpbox\captionbox
\usepackage{natbib}

\begin{document}
\twocolumn[

\icmltitle{The Expressive Power of Word Embeddings}

\icmlauthor{Yanqing Chen
}{cyanqing@cs.stonybrook.edu}
\icmlauthor{Bryan Perozzi
}{bperozzi@cs.stonybrook.edu}
\icmlauthor{Rami Al-Rfou'
}{ralfrou@cs.stonybrook.edu}
\icmlauthor{Steven Skiena}{skiena@cs.stonybrook.edu}
\icmladdress{Computer Science Dept. Stony Brook University
             Stony Brook, NY 11794}

\icmlkeywords{machine learning, word representation, ICML}

\vskip 0.3in
]

\begin{abstract}

We seek to better understand the information encoded in word embeddings. We propose several tasks that help to distinguish the characteristics of different publicly released embeddings. Our evaluation shows that embeddings are able to capture surprisingly nuanced semantics even in the absence of sentence structure.
Moreover, benchmarking the embeddings shows great variance in quality and
characteristics of the semantics captured by the tested embeddings. Finally, we
show the impact of varying the number of dimensions and the resolution of each
dimension on the effective useful features captured by the embedding space. Our
contributions highlight the importance of embeddings for NLP tasks and the effect of
their quality on the final results.

\end{abstract}

\section{Introduction}

Distributed word representations (embeddings) capture semantic and syntactic features of
words out of raw text corpus without human intervention or language dependent
processing. The features embedding capture are task independent which make them ideal for language modeling. However,
embeddings are hard to interpret and understand. Despite the efforts of visualizing the word embeddings \cite{van2008visualizing}, points in high dimensional spaces carry a lot of information that is hard to quantify. Additionally,  publicly available embeddings generated by multiple research groups use different data and training procedures and there is not yet an understanding about the best way to learn these representations. 




In this paper, we investigate four public released word embeddings: (1) HLBL, (2) SENNA, (3) Turian's and (4) Huang's. We use context-free classification tasks rather than sequence labeling tasks (such as part of speech tagging) to isolate the effects of context in making decisions and eliminate the complexity of the learning methods. Specifically, our work makes the following contributions:
\begin{itemize}

\item We show through evaluation that embeddings are able to capture semantics in the absence of sentence structure and that there is a difference in the characteristics of the publicly released word embeddings.

\item We explore the impact of the number of dimensions and the resolution of
each dimension on the quality of the information that can be encoded in the
embeddings space. That shows that minimum effective space needed to capture the
useful information in the embeddings.

\item We demonstrate the importance of word pair orientation in encoding useful
linguistic information. We run two pair classification tasks and provide an example with one of them
where pair performance greatly exceeds that of individual words.
\end{itemize}

The rest of the work proceeds as follows: First we describe the word embeddings we consider.  Next we discuss our classification experiments, and present their results. Finally we discuss the effects of scaling down the size of the embeddings space.

\section{Related Work}

The original work for generating word embeddings was presented by Bengio et.\
al.\ in \cite{bengio2003neural}.   The embeddings were a secondary output when generating language model.
Since \cite{bengio2003neural}, there has been a significant interest in speeding up the generation process \cite{bengio2003quick, bengio2009curriculum}. 
These original language models were evaluated using perplexity. We argue that while perplexity 
is a good metric of language modeling, 
it is not insightful about how well the embeddings capture diverse types of information. 

SENNA's embeddings \cite{collobert2011deep} are generated using a model that is discriminating and non-probabilistic. In each training update, we read an n-gram from the corpus, concatenating the learned embeddings of these n words. Then a corrupted n-gram is used by replacing the word in the middle with a random one from the vocabulary. On top of the two phrases, the model learns a scoring function that scores the original phrases lower than the corrupted one. The loss function used for training is hinge loss. \cite{collobert2011natural} shows that embeddings are able to perform well on several NLP tasks in the absence of any other features.  The NLP tasks considered by SENNA all consist of sequence labeling, which imply that the model might learn from sequence dependencies. Our work enriches the discussion by focusing on term classification problems. 

In \cite{turian2010word}, Turian et.\ al.\ duplicated the SENNA embeddings with some differences; they corrupt the last word of each n-gram instead of the word in the middle. They also show that using embeddings in conjunction with typical NLP features improves the performance on the Named Entity Recognition task. An additional result of \cite{turian2010word} shows that most of the embeddings have similar effect when added to an existing NLP task.  This gives the wrong impression. Our work illustrates that not all embeddings are created equal and there are significant differences in the information captured by each publicly released model exist.

Mnih and Hinton \cite{mnih2007three} proposed a log-bilinear loss function to model language. Given an n-gram, the model concatenates the embeddings of the n-1 first words, and learns a linear model to predict the embedding of the last word.  Mnih and Hinton later proposed Hierarchical log-bilinear (HLBL) model embeddings \cite{mnih2009scalable} to speed up model evaluation during training and testing by using a hierarchical approach (similar to \cite{morin2005hierarchical}) that prune the search space for the next word by dividing the prediction into a series of predictions that filter region of the space. The language model eventually is evaluate using perplexity.

A fundamental challenge for neural language models involves representing words which have multiple meanings.  In \cite{huang2012improving}, Huang et.\ al.\ incorporate global context to deal with challenges raised by words with multiple meanings.

Recent work by Mikolov et.\ al.\ \cite{mikolov2013linguistic} investigates linguistic regularities captured by the relative positions of points in the embedding space.  Our results regarding pair classification are complementary.
\section{Experimental setup}

We will construct three term classification problems and two
pair classification problems to quantify the quality of the embeddings.

\subsection{Evaluation Tasks}
Our evaluation tasks are as follows:

\begin{itemize}

\item \textbf{Sentiment Polarity}: We use Lydia's sentiment lexicon \cite{godbole2007large} to create sets of words which have positive or negative connotations and construct the 2-class sentiment polarity test. The data size is 6923 words.

\item \textbf{Noun Gender}: We use Bergsma's dataset \cite{Bergsma:06} to compile a list of masculine and feminine proper nouns. Names that co-refer more frequently with \emph{she}/\emph{he} are respectively considered feminine/masculine. Strings that co-refer the most with \emph{it}, appear less than 300 times in the corpus, or consist of multiple words are ignored. The total size is 2133 words.

\item \textbf{Plurality}: We use WordNet \cite{fellbaum2010wordnet} to extract
nouns in their singular and plural forms. The data consists of 3012 words.

\item \textbf{Synonyms and Antonyms}: We use WordNet to extract synonym and
antonym pairs and check whether we can part one kind from the others. The relation is symmetric thus we put each word pair together with their order-reversed-counterparts. There are 3446 different word pairs.


\item \textbf{Regional Spellings}: We collect the words that differ in spelling
between UK English and the American counterpart from an online source
\cite{WWL:2009:Online}. We make this task be a pair classification task to emphasize relative distances between embeddings. We have 1565 pairs in this task.

\end{itemize}

We ensure that for all tasks the class labels are balanced. This allow our baseline evaluation to be either the random classifier or the most frequent label classifier. Either of them will give an accuracy of 50\%. Table \ref{examples} shows examples of each of the 2-class evaluation tasks. The classifier is asked to identify which of the classes a term or pair belongs to.
\begin{table}[t]
\vskip 0.15in
\begin{scriptsize}
\setlength{\tabcolsep}{1pt}
\begin{subfigure}[t]{\textwidth}
\begin{tabular}{|c|ll|ll|ll|}
\hline
& \multicolumn{2}{c|}{\textbf{Sentiment}} & \multicolumn{2}{c|}{\textbf{Noun Gender}} & \multicolumn{2}{c|}{\textbf{Plurality}} \\
\multirow{2}{*}{\ } & Positive & Negative & Feminine & Masculine & Plural & Singular\\
\cline{1-7}
\multirow{3}{*}{Samples} &good & bad & Ada & Steve & cats & cat\\
&talent & stupid & Irena & Roland & tables & table\\
&amazing & flaw & Linda & Leonardo& systems & system\\
\hline
\end{tabular}
\end{subfigure}
\begin{subfigure}[b]{\textwidth}
\begin{tabular}{|c|ll|ll|}
\hline
& \multicolumn{2}{c|}{\textbf{Synonyms and Antonyms}} & \multicolumn{2}{c|}{\textbf{Regional Spellings}} \\
\multirow{2}{*}{\ } & Synonyms & Antonyms & UK & US \\
\cline{1-5}
\multirow{3}{*}{Samples} & store shop & rear front & colour & color\\
& virgin pure & polite impolite & driveable & drivable\\
& permit license & friend foe & smash-up & smashup\\
\hline
\end{tabular}
\end{subfigure}
\end{scriptsize}
\caption{Example input from each task}
\label{examples}
\vskip -0.1in
\end{table}

\subsection{Embeddings' Datasets}
We choose the following publicly available embeddings datasets for evaluation.  

\begin{itemize}
\item \textbf{SENNA's embeddings} covers 130,000 words with 50 dimensions for
each word. 
\item \textbf{Turian's embeddings} covers 268,810 words, each represented either with 25, 50 or 100 dimensions.  

\item \textbf{HLBL's embeddings} covers 246,122 words.  These embeddings were trained on same data used for Turian embedding for 100
epochs (7 days), and have been induced in 50 or 100 dimensions.

\item \textbf{Huang's embeddings} covers 100,232 words, in 50 dimensions.  Huang's embeddings require context to disambiguate which prototype to use for a word. Our tasks are context free so we average the multiple prototypes to a single point in the space. (This was the approach which worked best in our testing.)

\end{itemize}

It should be emphasized that each of these models has been induced under substantially different training parameters.  Each model has its own vocabulary, used a different context size, and was trained for a different number of epochs on its training set. While the control of these variables is outside the scope of this study, we hope to mitigate one of these challenges by running our experiments on the vocabulary shared by all these embeddings. The size of this shared vocabulary is 58,411 words.


\subsection{Classification}
For classification we used Logistic Regression and a SVM with the RBF-kernel as linear and non-linear classifiers. There is a model-selection procedure by running a grid-search on the parameter space with the help of the development data. All experiments were written using the Python package Scikit-Learn \cite{scikit-learn}. For the term classification tasks we offered the classifier only the embedding of the word as an input. For pairwise experiments, the input consists of the embeddings of the two words concatenated. 

The average of four folds of cross validation is used to evaluate the
performance of each classifier on each task. 50\%, 25\%, 25\% of the
data are used, as training, development and testing datasets respectively, for
evaluation and model selection.

\section{Evaluation Results}
Here we present the evaluation of both our term and pair classification results.

\subsection{Term Classification}
\label{TermClassification}

\begin{figure}[htb]
	 \centering
     \includegraphics[width=0.45\textwidth]{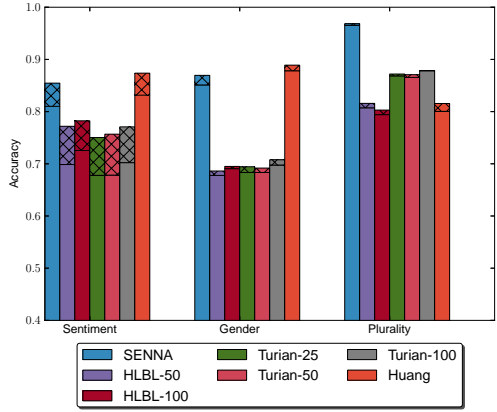}
  	    ~ 
     \caption{Results of the term-based tasks considered, shaded areas represent improvements using kernel SVM.}
     \label{fig:term2}
\end{figure}

Figure \ref{fig:term2} shows the results over all the 2-class term classification tasks using logistic regression and RBF-kernel SVM.  It is surprising that all the embeddings we considered did much better than the baseline, even on a seemingly hard tests like sentiment detection. What's more, there is strong performance from both the SENNA and Huang embeddings.  SENNA embeddings seem to capture the plurality relationship better, which may be from the emphasis that the SENNA embeddings place on shallow syntactic features.



Table \ref{strongweak} shows examples of words from the test datasets after classifying them using logistic regression on the SENNA embeddings. The top and bottom rows show the words that the classifier is confident classifying, while the rows in the middle show the words that lie close to the decision boundary. For example, \emph{resilient} could have positive and negative connotations in text, therefore, we find it close to the region were the words are more neutral than being polarized.

For SENNA, the best performing task was the Plurality task. That explains the obvious contrast between the probabilities given to the words. The top words are given almost 100\% probability and the bottom ones are given almost 0\%. The results of regional spelling task is shown here in the term-wise setup. Despite not performing as well as the pair-wise spelling, we can see that classifier shows meaningful results. We can clearly notice that the British spellings of words favor the usage of hyphens, \emph{s} over \emph{z} and \emph{ll} over \emph{l}.

\begin{table}[!htb]
\begin{center}
\begin{adjustwidth}{-0.3cm}{}
\begin{scriptsize}
\begin{subfigure}[b]{\textwidth}
\begin{tabular}{l|l|r|l|l|r|}
\cline{2-3}\cline{5-6}
& \textbf{Positive} & Prob &   & \textbf{British} & Prob\\
\cline{2-3}\cline{5-6}
\multirow{16}{2mm}{\begin{sideways}\textbf{Sentiment}\end{sideways}}
& world-famous & 99.85 &
\multirow{16}{2mm}{\begin{sideways}\textbf{Regional Spelling}\end{sideways}}& kick-off & 92.37 \\

& award-winning & 99.83 & &hauliers&91.54 \\
& high-quality & 99.83 & &re-exported&89.46 \\
& achievement & 99.81 & &bullet-proof&88.69 \\
& athletic & 99.81 & &initialled&88.42 \\ 
\cline{2-3}\cline{5-6}

& resilient & 50.14 & &paralysed&50.16 \\
& ragged & 50.11 & &italicized&50.04 \\
& discriminating & 50.10 &&exorcise&50.03 \\

& stout & 49.97 &&fusing&49.90 \\
& lose & 49.83 &&lacklustre&49.78 \\
& bored & 49.81 &&subsidizing&49.77 \\
\cline{2-3}\cline{5-6}
 & bloodshed & 0.74 &&signaling&32.04 \\
& burglary & 0.68 &&hemorrhagic&21.69 \\
& robbery & 0.58 &&tumor&21.69 \\
& panic & 0.45  &&homologue&19.53 \\
& stone-throwing & 0.28 &&localize&17.50 \\
\cline{2-3}\cline{5-6}

& \textbf{Negative} & 1.0-Prob&& \textbf{American} & 1.0-Prob \\
\cline{2-3}\cline{5-6}
\end{tabular}
\end{subfigure}

\end{scriptsize}
\end{adjustwidth}
\caption{Examples of the results of the logistic regression classifier on different tasks.}
\label{strongweak}
\end{center}
\end{table}

\subsection{Pair Classification}
\begin{figure}[!tbh]
        \begin{subfigure}[b]{0.45\textwidth}
    \includegraphics[width=\textwidth]{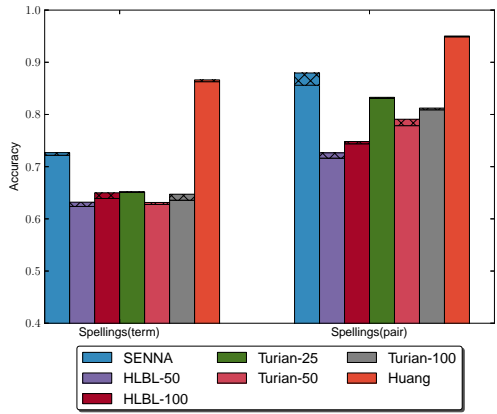}
    \caption{UK/US term vs. pair}
    \label{fig:ukusukus}
        \end{subfigure}%
        ~ 
        \newline
        \begin{subfigure}[b]{0.45\textwidth}
    			\includegraphics[width=\textwidth]{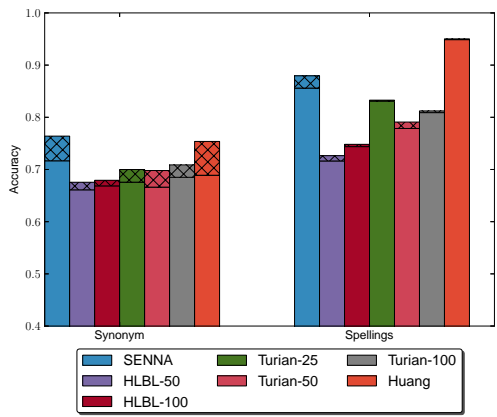}
    			\caption{2-class pair results}
  			\label{fig:pair2}
        \end{subfigure}
        ~ 
        \caption{Results of the pair-based tests.  Figure \ref{fig:ukusukus} shows the difference between treating the UK/US spellings as a single word problem, or using a pair of embeddings.   Figure \ref{fig:pair2} shows the results of the 2-class pair tests,  shaded areas represent improvements using kernel SVM.}
\label{fig:pair_results}
\end{figure}

Sometimes however, the choice to use pair classification can make quite a difference in the results. Figure \ref{fig:ukusukus} shows that classifying individual words according to their regional usage performs poorly. We can redefine the problem such that the classifier is asked to decide if the first word, in a pair of words, is the American spelling or not. Figure \ref{fig:ukusukus} shows that performance improves a lot. This hints that the words under this criteria are not separable by a hyper-plane in any subspace of the original embeddings space. Instead, we draw a similar conclusion as \cite{mikolov2013linguistic} that the pairs' positions relative to each other is what encodes such information but not their absolute coordinates, and relationship between words often indicate the relative difference vector between corresponding points. 

In our previous Plurality test, the SENNA embeddings significantly outperformed Huang's.  However in our regional spelling task (which might seem similar), Huang's embeddings outperform SENNA in both term and pair classification setups.  We believe that Huang's approach for building word prototypes from significant differences in context provide a significant advantage on this task.

We note that it is surprising that neural language models may capture the relation between a synonym and antonym.  Both the language modeling of HLBL and the way that SENNA/Turian corrupted their examples favor words that can syntactically replace each other; e.g. \emph{bad} can replace \emph{good} as easily as \emph{excellent} can.  The result of this syntactic interchangeability is that both \emph{bad} and \emph{excellent} are close to \emph{good} in the embedding space.  



\section{Information reduction}
Distributed word representation exist in continuous space, which is quite different from common language modeling techniques.
Beside the powerful expressiveness that we demonstrated previously, another key advantage of distributed representations is their size -  they require far less memory and disk storage than other techniques.  In this section we seek to understand exactly how much space word embeddings need in order to serve as useful features. We also investigate whether the powerful representation that embeddings offer is a result of having real value coordinates or the exponential number of regions which can be described using multiple independent dimensions. 


\subsection{Bitwise Truncation}
To reduce the resolution of the real numbers that make up the embeddings matrix. First we scale them to 32 bit integer values, then we divide the values by $2^{b}$, where $b$ is the number of bits we wish to remove. Finally, we scale the values back to lie between $(-1, 1)$. After this preprocessing we give the new values as features to our classifiers. In the extreme case, when we truncate 31 bits, the values will be all either $\{1, -1\}$.

\begin{figure}[htb]
    \includegraphics[width=0.45\textwidth]{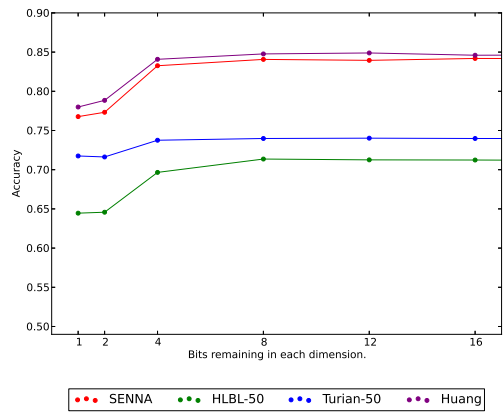}
    \centering
    \caption{Results of reducing the precision of the embeddings, averaged by the geometric mean of classifiers. We note that after removing 31 bits, each dimension of the embeddings is a binary feature.}
    \label{fig:prec_emb}
\end{figure}

Figure \ref{fig:prec_emb} shows that when we remove 31 bits (i.e, values are $\{1, -1\}$), the performance of an embedding dataset drops no more than 7\%. This reduced resolution is equivalent to $2^{50}$ regions which can be encoded in the new space.  This is still a huge resolution, but surprisingly seems to be sufficient at solving the tasks we proposed.  A na\"{\i}ve  approximation of this trick which may be of interest is to simply take the the sign of the embedding values as the representation of the embeddings themselves.

\subsection{Principle Component Analysis}
\begin{figure}[htb]
    \includegraphics[width=0.45\textwidth]{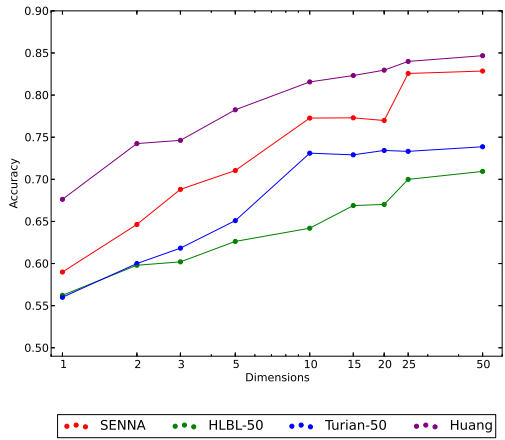}
    \centering
    \caption{Results of reducing the dimensions of the embeddings through PCA, averaged by the geometric mean.}
    \label{fig:pca_emb}
\end{figure}

The bitwise truncation experiment indicates that the number of dimensions could be a key factor into the performance of the embeddings.  To experiment on this further, we run PCA over the embeddings datasets to evaluate task performance on a reduced number of dimensions. Figure \ref{fig:pca_emb} shows that reducing the dimensions drops the accuracy of the classifiers significantly across all embedding datasets. 



Another key difference between the truncation experiment and the PCA experiment is that the truncation experiment may preserve relationships captured by non-linearities in the embedding space.  Linear PCA can not offer such guarantees and this weakness may contribute to the difference in performance.  

\section{Conclusion}
Distributed word representations show a lot of promise to improve supervised learning and semi-supervised learning. The practical advantages of having dense representations make them ideal for industrial applications and software development. The previous work mainly focused on speeding up the training process with one metric for evaluation, perplexity. We show that this metric is not able to provide a nuanced view of their quality. We develop a suite of linguistic oriented tasks which might serve as a part of a comprehensive benchmark for word embedding evaluation. The tasks focus on words or pairs of them in isolation to the actual text.  The goal here is not to build a useful classifier as much as it is to understand how much supervised learning can benefit from the features which are encoded in the embeddings.

We succeed in showing that the publicly available datasets differ in their quality and usefulness, and our results are consistent across tasks and classifiers. Our future work will try to address the factors that lead to such diverse quality. The effect of training corpus size and the choice of the objective functions are two main areas where better understanding is needed. 

While our tasks are simple, the differences among task performance shed light on the features encoded by embeddings. We showed that in addition to the shallow syntactic features like plural and gender agreement, there are significant semantic partitions regarding sentiment and synonym/antonym meaning. Our current tasks focus on nouns and adjectives, and the suite of tasks has to be extended to include tasks that address verbs and other parts of speech. 
\section*{Acknowledgments}
This research was partially supported by NSF Grants DBI-1060572 and
IIS-1017181, with additional support from TexelTek, Inc.

\bibliography{myrefs}
\bibliographystyle{plainnat}
\clearpage
\section *{Supplemental Materials}
\markright{SUPPLEMENTAL MATERIALS}

\subsection*{3-class Tests}
\begin{figure}[!htb]
                
    \includegraphics[width=0.5\textwidth]{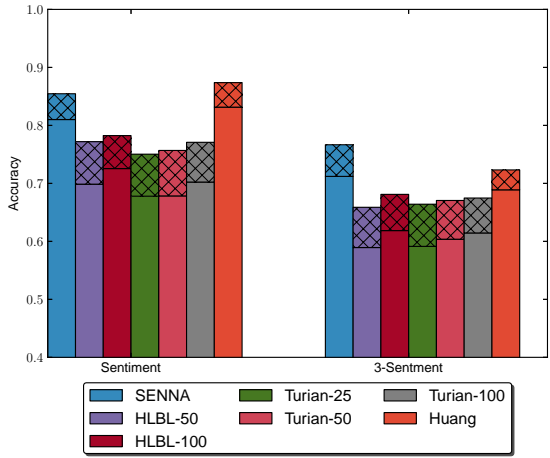}
    \caption{The performance on the 3-class version of the sentiment task,  shaded areas represent improvements using kernel SVM.} 

   \label{fig:term3}
\end{figure}

To strengthen these results, we performed a 3-class version of the sentiment test, in which we evaluated the ability to classify words as having positive, negative, or neutral sentiment value.  The results are presented in Figure \ref{fig:term3}.  The results are consistent with those from our 2-label test, and all embeddings perform much higher than the baseline score of 33\%.  


\begin{figure}[!htb]
                
    \includegraphics[width=0.5\textwidth]{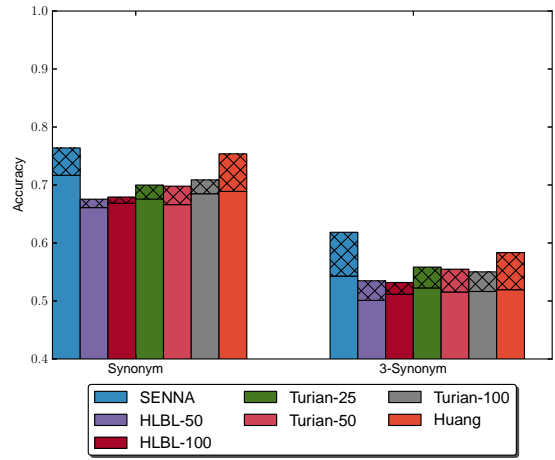}
    \caption{The performance of the 3-class synonym/antonym task,   shaded areas represent improvements using kernel SVM.}
    \label{fig:pair3}
\end{figure}

In order to investigate the depth to which synonyms and antonyms are captured, we conducted a 3-class version of the same test.  We now evaluate between pairs of words that are synonyms, antonyms, or  have no such relation.  While such a task is much harder for the embeddings, the results in Figure \ref{fig:pair3} show that a nonlinear classifier can capture the relationship, particularly with the SENNA embeddings.  An analysis of the confusion matrix for the nonlinear SVM showed that errors occurred roughly evenly between the classes. We believe that this finding regarding the encoding of synonym/antonym relationships is an interesting contribution of our work.

\subsection*{Dimensional Reduction by task}

\begin{figure}[!h]
    \includegraphics[width=0.5\textwidth]{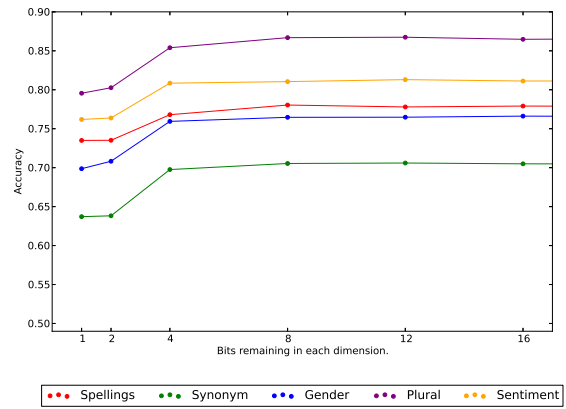}
    \caption{Results of reducing the precision of the embeddings, averaged by the geometric mean of classifiers acrossing tasks}
    \label{fig:prec_task}
\end{figure}

\begin{figure*}[!ht]
	\centering
	\begin{subfigure}[b]{0.49\textwidth}   
	\centering
    \includegraphics[width=\textwidth]{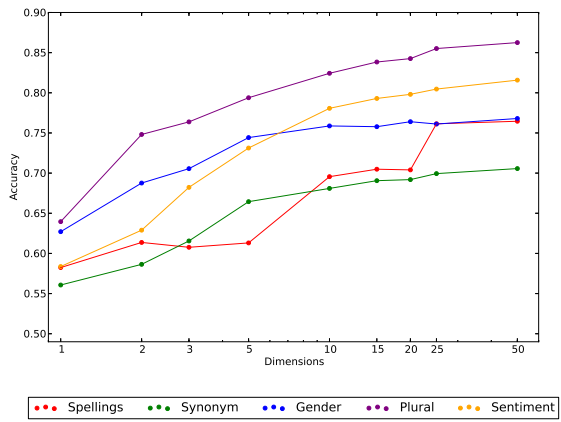}
    \caption{By task}
    \label{fig:pca_task}
   	\end{subfigure}
	\begin{subfigure}[b]{0.49\textwidth}   
	\centering	
    \includegraphics[width=\textwidth]{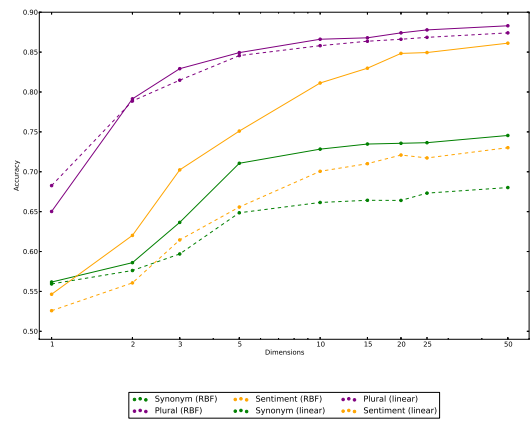}
    \caption{Linear vs. Nonlinear}
    \label{fig:pca_nonlinear}
   	\end{subfigure}   	
    \caption{Results of reducing the dimensions of the embeddings through PCA, averaged by the geometric mean across tasks (\ref{fig:pca_task}).  Figure \ref{fig:pca_nonlinear} shows the difference between linear (dashed) and non-linear (solid) classifiers for  our harder tasks (sentiment and synonym) and an easy task (plural).  The performance of the linear and nonlinear classifiers converges as PCA removes more dimensions.  This results in significantly degraded performance on nuanced tasks like sentiment analysis.}
    \label{fig:pca}  	   	
\end{figure*}

Looking at Figure \ref{fig:pca_task}, reducing the words embeddings to points on a real line almost deletes all the features that are relevant to the pair classification and to less a degree the sentiment features. Despite the 10\%-20\% drop in accuracy in the Plurality and Gender tasks, the classification is still higher than random. The results show that when  that shallow syntactic features such as gender and number agreement are preserved at the expense of more subtle semantic features such as sentiment polarity.  This gives us insight into what the hierarchical structure of the embeddings space looks like.  Shallow semantic features are present in all aspects of the space, and when PCA chooses to maximize this variance of the feature space it is at the expense of the other semantic properties.

We also illustrate this phenomenon in Figure \ref{fig:pca_nonlinear}, by showing how the performance of the linear and non-linear classifiers converge for our harder tasks (sentiment and synonym) as we reduce the number of dimensions with PCA.

\end{document}